\documentclass[conference]{IEEEtran}
\IEEEoverridecommandlockouts
\usepackage{cite}
\usepackage{amsmath,amssymb,amsfonts}
\usepackage{algorithmic}
\usepackage{graphicx}
\usepackage{textcomp}
\usepackage{xcolor}
\usepackage{booktabs}
\usepackage{multirow}
\usepackage{seqsplit}

\usepackage{url}
\usepackage{graphicx}
\def\BibTeX{{\rm B\kern-.05em{\sc i\kern-.025em b}\kern-.08em
    T\kern-.1667em\lower.7ex\hbox{E}\kern-.125emX}}
\usepackage{hyperref}
\hypersetup{
    colorlinks=true,
    allcolors=blue,
    pdftitle={Process Mining-Enabled Federated Anomaly Detection},
    pdfpagemode=FullScreen,
}

\usepackage{pbalance}

\makeatletter
\newcommand{\linebreakand}{%
  \end{@IEEEauthorhalign}
  \hfill\mbox{}\par
  \mbox{}\hfill\begin{@IEEEauthorhalign}
}
\makeatother

\newif\ifblind
\blindfalse   
    
\begin{document}

\title{Process Mining-Enabled Federated Anomaly Detection for the Industrial IoT
\thanks{
The work of Francesco Vitale was partly supported by the PRIN 2022 project “MEDITATE: Towards Model-based Generation and Optimization of Advanced Driver Assistance Systems Testing Scenarios in Co-simulation” (CUP: E53C2400244006), funded by the Italian Ministry of University and Research (MUR). The work was also supported in part by a research grant from the Telecommunications Advancement Foundation (TAF), Japan. Views and opinions are those of the authors only and do not necessarily reflect those of the funding agencies, which cannot be held responsible for them.
}
}

\ifblind
\author{%
\IEEEauthorblockN{Anonymous Authors}
\IEEEauthorblockA{%
Affiliations withheld for double-blind review.}
}
\else
\author{%
\IEEEauthorblockN{Francesco Vitale}
\IEEEauthorblockA{
\textit{DIETI}\\
\textit{University of Naples Federico II}\\
Naples, Italy\\
francesco.vitale@unina.it}
\and
\IEEEauthorblockN{Salvatore Bramante}
\IEEEauthorblockA{
\textit{IMT School for Advanced Studies Lucca}\\
Lucca, Italy\\
salvatore.bramante@imtlucca.it}

\linebreakand

\IEEEauthorblockN{Francesco Flammini}
\IEEEauthorblockA{
\textit{DIMAI}\\
University of Florence\\
Florence, Italy\\
francesco.flammini@unifi.it}
\and
\IEEEauthorblockN{Hangli Ge}
\IEEEauthorblockA{
\textit{Interfaculty Initiative in Information Studies}\\
\textit{The University of Tokyo}\\
Tokyo, Japan\\
hanglige@g.ecc.u-tokyo.ac.jp}
}
\fi

\maketitle

\begin{abstract}
While federated anomaly detection enables the integration of local data insights across diverse nodes to reduce false positives, it inherently struggles with non-IID data. In practice, industrial IoT plants often operate under varying conditions and experience diverse environmental phenomena, resulting in significant heterogeneity in the monitored time series. In such settings, standard federated learning approaches fail to perform nuanced aggregation that accounts for these cross-node differences, ultimately leading to increased false positive rates. In this paper, we propose a process mining approach to evaluate the similarity of local node data against a reference provided by a global node. The global node utilizes these similarity metrics to dynamically weight local models during aggregation, producing personalized updates distributed across the network. We evaluate our approach using industrial IoT time-series data collected from manufacturing plants operating under distinct conditions. Experimental results demonstrate that our process mining-enabled federated anomaly detection pipeline achieves superior performance, yielding a 46.52\% and 11.99\% relative improvement in F1-score compared to isolated local training baselines and standard FedAvg, respectively, alongside a 51.24\% relative reduction in false positives over standard federated aggregation.
\end{abstract}


\section{Introduction}
\label{sec:introduction}

Anomaly detection for industrial IoT is an ever-popular task that allows evaluating whether the monitored plants are operating as expected and diagnosing faults that may occur due to physical disruptions or plant wear~\cite{chatterjee2022iotad}. Typically, industrial IoT applications involve multiple plants working in different operating conditions that may alternate over time~\cite{belay2023unsupervisedad}. Notably, this makes the ability to characterize normal behavior more challenging, as modern anomaly detection approaches involve machine learning algorithms that learn normal behavior from data and may struggle with either over-generalization or overfitting~\cite{de2022deepgenai}. 

To account for the distributed nature of industrial IoT applications, federated anomaly detection can be used. This paradigm involves aggregating diverse knowledge from different local nodes to improve the ability to discriminate normal behavior from anomalous observations~\cite{dong2025fadngs}. This paradigm is advantageous in distributed environments thanks to the reduction in communication overhead of local data to a centralized server and the increased privacy guarantees. However, a critical issue plagues federated anomaly detection, namely non-IID data~\cite{sattler2020flnoniid}. As mentioned, this issue is frequently encountered in industrial IoT plants, where varying operating conditions make it difficult to establish a solid reference normal profile and accurately evaluate anomalous behavior. 

In order to deal with non-IID data, several federated anomaly detection approaches have been proposed. They perform client clustering based on node similarity from local models' weights~\cite{liu2024casa,fan2024rethinkingclusteredfl,laridi2024enhanced} or evaluate clients' similarity based on data distribution indices~\cite{zhang2024fedhc,dong2025fadngs,jimenezgutierrez2025fedlecc, ge2025xckan, ge2026extendingfairprinciplesobservability}. However, these approaches often incur significant computational and communication overhead during the federated rounds, or they struggle to accurately capture the complex, sequential operational variations specific to industrial processes. 

To overcome these limitations, we propose a process mining-enabled federated anomaly detection approach. Process mining is a research area able to automatically extract behavioral models from event data and verify conformance of new data against such models~\cite{aalst2022pmhb}; notably, process mining has been successfully applied to time-series sensor data~\cite{vitale2026pmfd,shi2024industrialpad,vitale2023pmbasedad}. By capturing the underlying behavioral control-flow of the industrial assets, process mining allows constructing a similarity matrix that provides a weight distribution that guides the personalized aggregation of the anomaly detection models. This naturally mitigates the non-IID problem while preserving communication efficiency in federated anomaly detection.

In summary, the contributions of this paper are:
\begin{itemize}
    \item Process mining-based similarity evaluation of the behavior of different local nodes collecting industrial IoT data;
    \item The integration of the process mining-based similarity evaluation in a comprehensive federated anomaly detection framework.
\end{itemize}

The rest of the paper is structured as follows. Section~\ref{sec:background} establishes the background on process mining for sensor data and federated anomaly detection. Section~\ref{sec:approach} describes the proposed process mining-enabled federated anomaly detection approach. Section~\ref{sec:experiments} presents the preliminary experiments. Section~\ref{sec:conclusions} concludes and reports future works.
\section{Related Work}
\label{sec:background}
This section establishes the related work on process mining for sensor data and federated anomaly detection.

\subsection{Process Mining for IoT Data}
Process mining involves two main objects: event logs and process models~\cite{aalst2022pmhb}. An event log can be defined as a multiset\footnote{A multiset is a set with possibly repeated elements.} of traces of ordered events, where an event may indicate an activity of a business process or state transitions in an industrial plant. A process model is a representation of the reference application's behavior using a modeling notation that reports the dynamic relationships between the modeled events. The two main types of process mining leveraging these objects are process discovery, a set of algorithms dedicated to the automatic modeling of event relationships, and conformance checking, a set of algorithms that verify the compliance of new data against a reference process model~\cite{aalst2022pmhb}. 

Although initially used for modeling and optimizing business processes, process mining is increasingly applied to industrial operations. Driven by the pervasiveness of IoT applications, the community now frequently targets low-level sensor data~\cite{brzychczy2025pmsensordata}. This class of data records the physical plant states from dedicated sensors and the actions performed by actuators to control the plant. Unlike discrete business events, data gathered from physical plants usually consists of numerical time series and low-level logs that capture the continuous evolution of environmental phenomena or mechanical movements performed by the plant itself~\cite{vaneck2016enablingpm}. Practical applications include industrial water processing and distribution~\cite{myers2018adics}, manufacturing systems~\cite{vitale2026pmfd}, smart building~\cite{kitagishi2022wi, ge2024time} and coal gasification~\cite{shi2024industrialpad}. Similar to these works, we employ unsupervised state-based characterization of time-series data collected from IoT plants to extract their behavioral profile and use conformance checking to evaluate the plant similarities.

\subsection{Federated Anomaly Detection}

Federated learning is a distributed machine learning paradigm in which multiple devices collaboratively train a shared global model without centralizing raw data. In the standard federated learning setting, a central server coordinates iterative training rounds: it distributes the current global model to a set of participating clients, each with its private dataset and model. It aggregates the locally updated parameters into a new global model. McMahan et al.~\cite{mcmahan2017federatedlearning} demonstrated that this iterative model averaging approach, termed FedAvg, can train deep networks on decentralized and unbalanced data while reducing communication rounds by 10-100$\times$ compared to synchronized SGD. 

While FedAvg is widely adopted, its robustness under non-IID conditions remains contested. System heterogeneity compounds data heterogeneity, causing mis-clustering and degraded training in clustered federated learning environments. To mitigate this, recent literature increasingly leverages specialized client clustering. For example, Jimenez-Gutierrez et al.~\cite{jimenezgutierrez2025fedlecc} propose FedLECC, which groups clients by label-distribution similarity and prioritizes nodes with higher local empirical loss to optimize communication.

Similarly, Zhang et al.~\cite{zhang2024fedhc} introduce FedHC, which clusters dimensionally-reduced client gradients to group similar nodes and improves generalization via distance-weighted inter-group aggregation. In Industrial IoT anomaly detection, spatial-temporal distribution shifts introduce further complexities. Fan et al.~\cite{fan2024rethinkingclusteredfl} address this with ClusterFLADS, using PCA on final-layer parameters for efficient clustering, and leveraging catastrophic forgetting metrics to dynamically assign test data to the most suitable cluster model.

Aligning anomaly definitions across non-IID nodes is also critical to prevent false and missing detections. Dong et al.~\cite{dong2025fadngs} align these definitions securely by sharing noisy global density estimations instead of raw data. Furthermore, because autoencoder-based detection relies heavily on accurate thresholds, Laridi et al.~\cite{laridi2024enhanced} show that aggregating local summary statistics (e.g., mean and variance of reconstruction errors) into a global threshold significantly outperforms isolated local thresholding.

\begin{figure*}[!t]
    \centering
    \includegraphics[width=\linewidth]{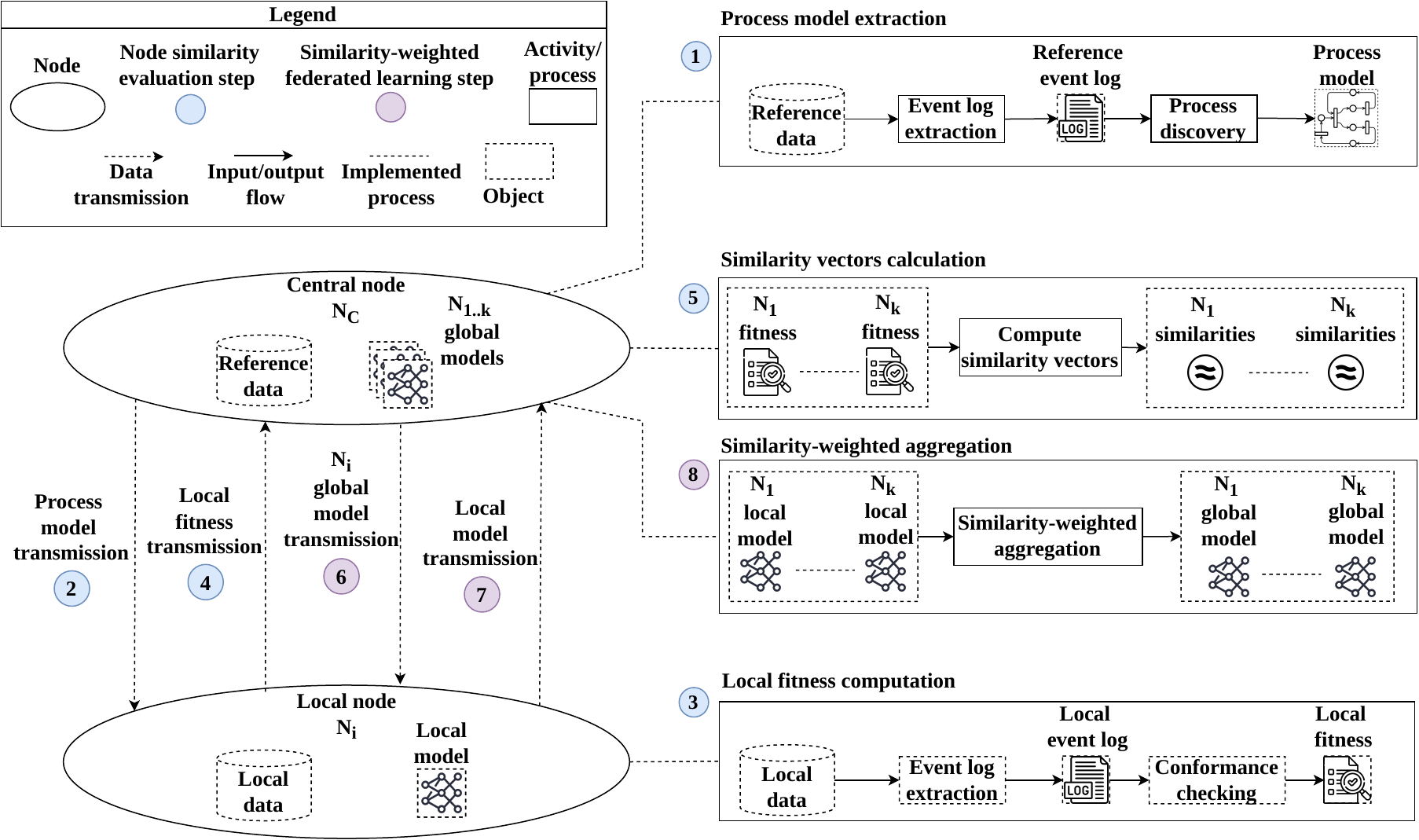}
    \caption{The proposed process mining-enabled federated anomaly detection approach: 1) the central node extracts a process model describing the behavior of reference data; 2) the process model is transmitted to a local node; 3) the local node computes the fitness between the received process model and its local data; 4) local fitness is transmitted to the central node; 5) the central node calculates the weights to assign during federated learning; 6) the central node transmits the selected global model for the given node; 7) the local node transmits its local model after a federated learning round; 8) the central node aggregates the models based on the previously computed weights.}
    \label{fig:approach}
\end{figure*}
While current methods successfully use gradient similarities or statistical summaries to stabilize federated training, they often overlook the sequential control-flow inherent to industrial processes. By extracting behavioral fingerprints directly from event logs via process mining, a structural similarity matrix can be built prior to training. This deterministic approach naturally guides the personalized aggregation of anomaly detection autoencoders, mitigating non-IID challenges while ensuring communication efficiency.

\section{The Proposed Approach}
\label{sec:approach}

In this section, we present the proposed process mining-enabled federated anomaly detection approach, represented in Fig.~\ref{fig:approach}. The approach can be decomposed into two consecutive parts: the node similarity evaluation through process mining and similarity-weighted federated learning. In the following description, we consider a distributed system consisting of a central node $N_C$ connected to $k$ local nodes $N_{1,\dots,k}$. Without loss of generality, we will focus our discussion on a general local node $N_{i}$.

\subsection{Node Similarity Evaluation}
The objective of this phase is to quantify the operational alignment between distributed nodes without requiring the exchange of raw sensor data. Instead of comparing local datasets directly, we leverage process mining techniques to evaluate how similarly each node's local behavior conforms to a shared, global process model. 

Initially, the central node performs \textbf{process model extraction} by utilizing historical or standardized reference data to extract a reference event log. The extraction of an event log from industrial IoT data can be performed through discretization of time series data through clustering techniques~\cite{vitale2023pmbasedad, vitale2026pmfd}. Specifically, the continuous, multi-dimensional sensor readings are first preprocessed and normalized to ensure uniform feature scaling. A clustering algorithm is then applied to partition the continuous state space into a predefined number of distinct clusters, $k$. Each resulting cluster centroid represents a unique, stable operational state of the industrial system. As widely documented in the literature, the number of clusters set for data discretization may influence the end results due to underfitting and overfitting phenomena related to process mining. 
When the reference event log is obtained, a process discovery algorithm can be used to generate a process model. The process model serves as the baseline representing the expected, standard operational behavior of the industrial IoT system. When the process model is available, it is transmitted to local nodes $N_{1,\dots,k}$.

At the local level, the node $N_{i}$ performs \textbf{local fitness computation} after receiving the process model from $N_C$. First, it proceeds to extract an event log from its own data using the same pre-processing parameters adopted by $N_C$. Instead of transmitting this data, the node performs conformance checking against the process model received from the central node. Conformance checking can also be performed with a plethora of methods. The most popular, used also for event logs extracted from time-series data, is alignment-based conformance checking, which attempts to align the actual behavior of the log to the behavior encoded in the process model~\cite{aalst2022pmhb}. The result of conformance checking is a local fitness score $F_{i}$ that quantifies how well the node's recorded behavior matches the global reference.

The local fitness scores are transmitted back to the central node to perform \textbf{similarity vectors calculation}. For any given reference node $N_{i}$ and another node $N_{j}$, the raw similarity $S_{i,j}$ is computed as the complement of the absolute difference between their fitness scores:

$$S_{i,j} = 1 - |F_{i} - F_{j}| \quad \text{for} \quad i \neq j$$

To account for a node's contribution to its own local model during the subsequent federated aggregation, the self-similarity is explicitly defined as a perfect match, i.e., $S_{i,i}=1.0$. While this first approach suits traditional conformance checking, as it deals with the main metric to assess process conformance, other types of process-related features can be extracted to assess the node similarities, such as alignment-based diagnoses \cite{vitale2025cfad}.

Finally, to ensure that the weights can be applied proportionally during federated learning, the raw similarity scores are normalized across all available nodes. The final normalized similarity weight $W_{i,j}$, representing the influence of node $N_{j}$ on the reference node $N_{i}$, is calculated as:

$$W_{i,j} = \frac{S_{i,j}}{\sum_{k=1}^{K} S_{i,k}}$$

\begin{figure*}[t!]
    \centering
    \includegraphics[width=\linewidth]{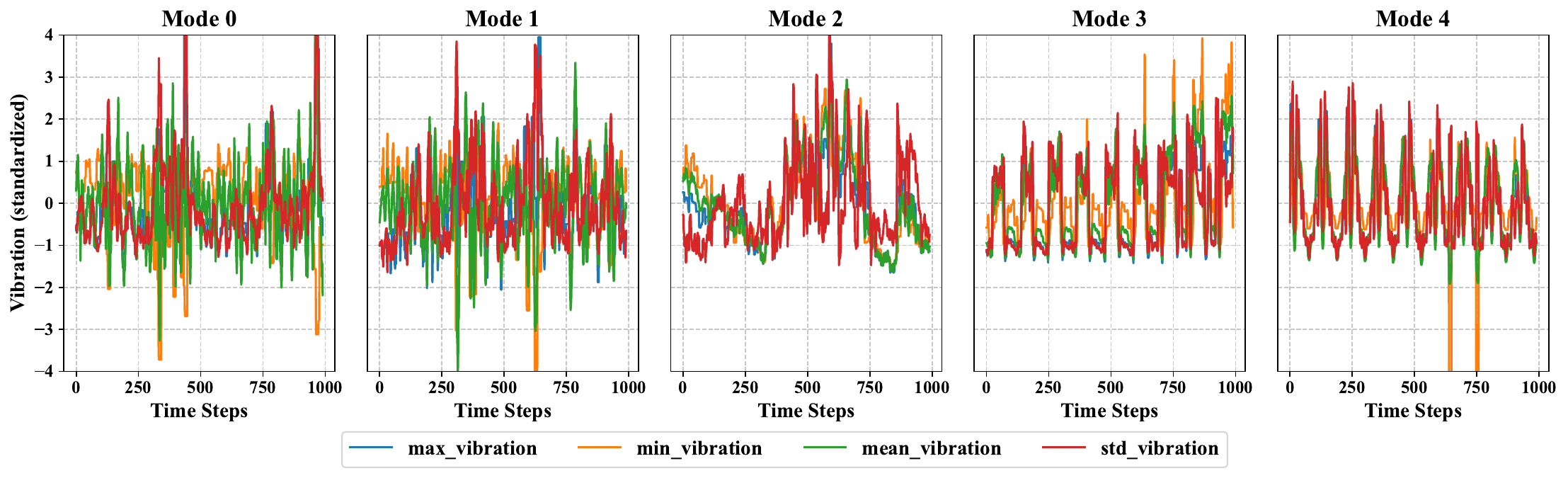}
    \caption{The synthetic features extracted from the vibration values of different operating modes of our dataset.}
    \label{fig:MODES_PLOT}
\end{figure*}
This computation yields a set of similarity vectors where the sum of weights for any target node equals exactly $1.0$. These vectors ensure that during the training phase, a local node will incorporate updates more heavily from peers that exhibit highly similar process behaviors, effectively mitigating the adverse effects of non-IID data distributions.

\subsection{Similarity-Weighted Federated Learning}
Building upon the similarity matrix derived from the process mining phase, the system executes a federated learning process to train the anomaly detection models. This phase consists of local training and similarity-weighted model aggregation.

During each communication round, every local node $N_{i}$ performs local model training using its locally extracted time-series features (e.g., maximum, minimum, mean, and standard deviation of vibration data). The objective of the model is to minimize a designated loss function, thereby learning the representations of normal operating data. Let $\theta_{i}^{t}$ denote the local model parameters (weights and biases) of node $N_{i}$ after local training at round $t$.

After local training, the nodes transmit their updated model parameters $\theta_{i}^{t}$ to the central server. Unlike standard FedAvg, which applies uniform or sample-size-based weighting, our approach involves \textbf{similarity-weighted aggregation}, utilizing the normalized process similarity weights $W_{i,j}$ computed in the previous phase. As illustrated in the architecture workflow, the central server calculates a customized, updated global model $\theta_{i}^{t+1}$ for each target node $N_{i}$ by aggregating the parameters of all participating nodes $N_{j}$ according to their behavioral similarity:

$$\theta_{i}^{t+1} = \sum_{j=1}^{K} W_{i,j} \theta_{j}^{t}$$

This similarity-weighted aggregation ensures that a node's updated model is primarily influenced by peers with structurally similar underlying processes. By preventing negative transfer from nodes with vastly different operational behaviors, this step effectively mitigates the performance degradation typically caused by non-IID data distributions across the federated network. 

Once the federated training converges, the optimized models are deployed for local anomaly detection. When a new data sequence is processed, the model evaluates it and generates a continuous anomaly score (e.g., a reconstruction or forecasting error).

\section{Experiments}
\label{sec:experiments}
In this section we evaluate the effectiveness of our process mining-enabled federated anomaly detection approach. We introduce the dataset used for our tests, the techniques adopted to implement the approach in Fig.~\ref{fig:approach}, the baselines, and discuss the results. The code implemented for dataset pre-processing and the implementation of the methods and baselines is available online\footnote{\url{https://github.com/francescovitale/pm_enabled_fad}.}.

\subsection{Dataset}
The dataset involves vibration data collected from a scale replica of an automotive manufacturing industry plant. The plant is a cart that transports car parts during its assembly, and is therefore subject to different load and wear conditions. We have monitored the data under five different conditions. Fig~\ref{fig:MODES_PLOT} shows the five operating modes, plotting four synthetic features obtained through sliding windows. Specifically, we utilize a sliding window of size 10 with a single-step overlap. Within each window, four statistical features are calculated: the maximum, minimum, mean, and standard deviation of the raw vibration signal. To ensure numerical stability and comparable scales across different operational states, the extracted features are subsequently subjected to standard scaling independently for each operating mode.

To simulate a distributed architecture, the five normalized operating modes are randomly shuffled and partitioned into distinct nodes. One operating mode is exclusively assigned as a global reference node, retaining the entirety of its standard-scaled data without further modification. The remaining four operating modes are distributed among four independent local nodes. Each local node receives the complete dataset of its exclusively assigned operating mode to serve as its local training set. To rigorously evaluate generalizability across different machine states, cross-node test datasets are constructed. For any given local node, its corresponding test set is obtained from another randomly selected local node. For example, the node with local training data of Mode 0 could be tested against the data of Modes 1 to 4.

To establish an evaluation ground truth for anomaly detection algorithms, subtle synthetic anomalies are injected into a randomly selected 15\% subset of each test dataset. An anomaly is generated by randomly selecting between one and three of the extracted features and shifting their normalized values by a random margin ranging from 1.0 to 2.5 standard deviations. Finally, each constructed test set is thoroughly shuffled to prevent temporal or sequential biases during model evaluation.

\subsection{Techniques and baselines}
The following is a description of the techniques and baselines used to compare our approach's performance. In all cases, the techniques and baselines are run 10 times to account for experimental noise.

\subsubsection{Our approach}
Regarding the node similarity evaluation, we have implemented k-means clustering to characterize different states from time-series data. The use of k-means is supported by prior work~\cite{vitale2026pmfd}, which demonstrated that a trade-off in process model quality and computational complexity can be achieved with different numbers of clusters. In this work, we selected a number of clusters equal to 3. Following this, to generate sound and perfectly fitting process models for the global node, we have selected the inductive miner. As for the similarity-weighted federated learning, we employ a symmetrical Deep Autoencoder trained in an unsupervised manner on nominal operational data. The encoder compresses the four-dimensional input ($d=4$) into a two-dimensional bottleneck via 8- and 4-neuron hidden layers with Rectified Linear Unit ($\text{ReLU}$) activations. The decoder symmetrically reconstructs the input through 4- and 8-neuron layers, concluding with a linear reconstruction layer. Models are optimized using Adam (learning rate of 0.001, batch size of 32) to minimize the Mean Squared Error ($\text{MSE}$) between the input $\mathbf{x} \in \mathbb{R}^4$ and its reconstruction $\hat{\mathbf{x}} \in \mathbb{R}^4$:
$$L_{\text{MSE}} = \frac{1}{4} \sum_{j=1}^{4} (x_j - \hat{x}_j)^2$$
Local updates execute for 25 epochs per round across 20 federated rounds, yielding 500 total epochs per node. During inference, the sample-wise reconstruction error serves as the anomaly score, since out-of-distribution patterns naturally yield higher errors. Binary classification thresholds are determined using the 95th percentile of the error on training data.

\subsubsection{Baselines}
The first baseline consists of an isolated, local-only training strategy. Under this regime, each local node trains an independent autoencoder exclusively on its own local dataset, without any communication, weight aggregation, or parameter sharing with other nodes. This strategy serves to measure the base capability of the network when it is entirely constrained by data silos and lacks access to generalized operational patterns.

The second baseline employs standard FedAvg, representing a traditional, homogeneous collaborative training approach. In this configuration, local models compute parameter updates on their respective datasets and transmit them to a central coordinating server. The server then aggregates these updates by computing a uniformly weighted average across all participating nodes. Specifically, each of the $N$ nodes is assigned an equal weight of $1/N$ during the aggregation step, regardless of underlying data heterogeneity or node similarity. The resulting globally averaged model is subsequently broadcast back to the local nodes to initialize the next round of training.

\subsection{Results}
\begin{table*}[htbp]
    \centering
    \caption{Anomaly detection performance across the baselines and the proposed similarity-weighted approach.}
    \label{tab:performance_metrics}
    \begin{tabular}{lccccc}
        \toprule
        \textbf{Strategy} & \textbf{F1-Score} & \textbf{TP (\%)} & \textbf{FP (\%)} & \textbf{TN (\%)} & \textbf{FN (\%)} \\
        \midrule
        No FL (Isolated) & $0.5692 \pm 0.0696$ & $10.3406 \pm 1.4579$ & $11.2185 \pm 4.1364$ & $73.8471 \pm 4.1364$ & $4.5938 \pm 1.4579$ \\
        Homogeneous FL (FedAvg) & $0.7447 \pm 0.1343$ & $11.9702 \pm 1.2895$ & $5.7013 \pm 4.1155$ & $79.3643 \pm 4.1155$ & $2.9642 \pm 1.2895$ \\
        Similarity-Weighted FL & $0.8340 \pm 0.0226$ & $12.6640 \pm 0.4503$ & $2.7800 \pm 0.8052$ & $82.2856 \pm 0.8052$ & $2.2704 \pm 0.4503$ \\
        \bottomrule
    \end{tabular}
\end{table*}
\begin{figure}[!t]
    \centering
    \includegraphics[width=0.85\columnwidth]{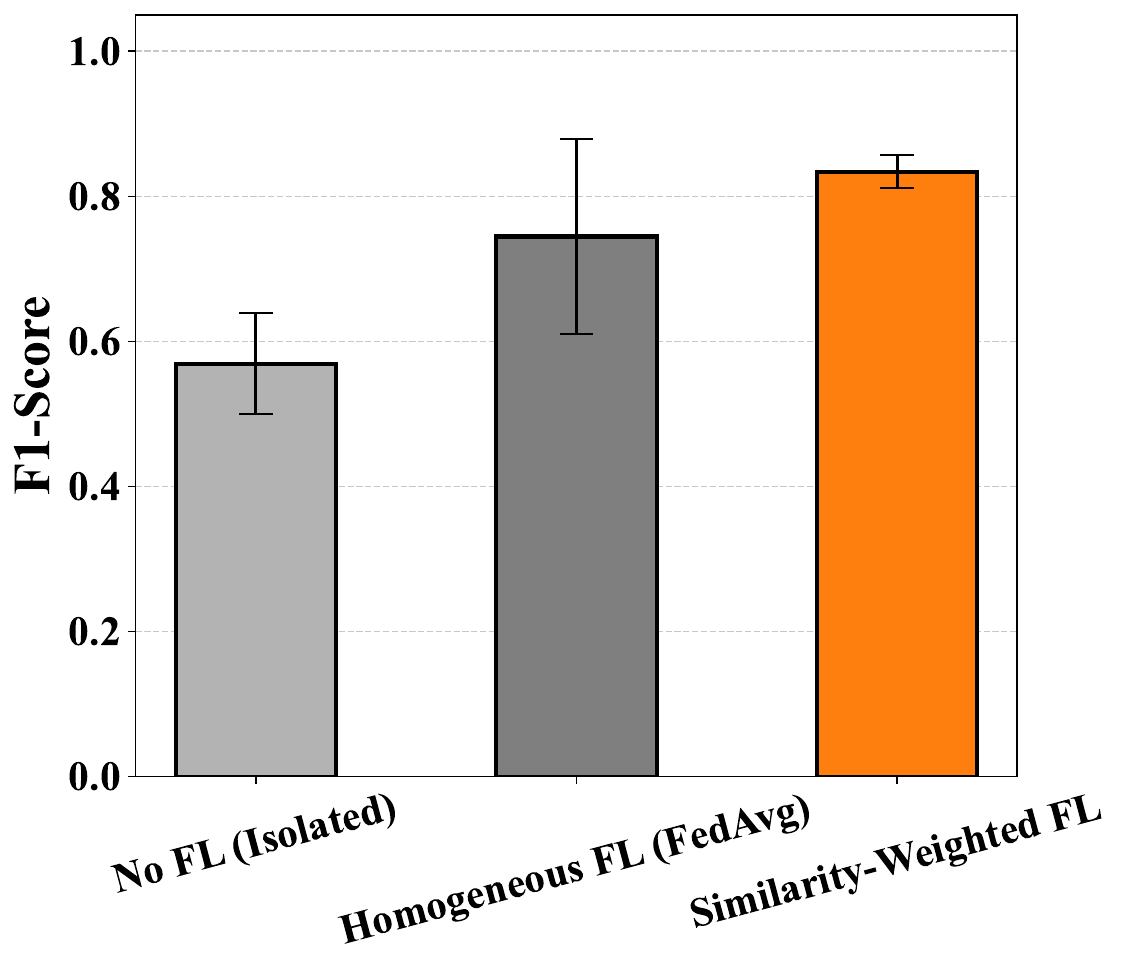}
    \caption{The comparison of F1 scores of the two baselines and the proposed similarity-weighted approach.}
    \label{fig:F1_PLOT}
\end{figure}
The experimental evaluation demonstrates the superior performance of the proposed process mining-enabled federated anomaly detection approach compared to standard baseline strategies. As illustrated in Table~\ref{tab:performance_metrics} and Figure~\ref{fig:F1_PLOT}, the similarity-weighted strategy achieves the highest average F1-Score of 0.8340 ± 0.0226, significantly outperforming both the isolated baseline (0.5692 ± 0.0696) and the FedAvg approach (0.7447 ± 0.1343). This translates to a relative F1-Score improvement of 46.52\% over the isolated method and 11.99\% over standard FedAvg. The isolated training strategy struggles due to its limited exposure to generalized operational patterns, while standard FedAvg suffers from performance degradation caused by non-IID data distributions across the federated network.  

Most notably, our method improves the reliability of the anomaly detection system by increasing the rate of true negatives and subsequently reducing false positives compared to standard federated approaches. According to the performance metrics, the similarity-weighted strategy yields a true negative rate of 82.2856\% ± 0.8052\% and restricts false positives to just 2.7800\% ± 0.8052\%. When compared directly to the standard FedAvg baseline—which recorded a 5.7013\% false positive rate and a 79.3643\% true negative rate—our approach achieves a 51.24\% relative reduction in false positives and a 3.68\% relative increase in true negatives. This improvement is a direct result of the process mining-driven weighting mechanism, which evaluates the structural similarity of underlying industrial processes before aggregation. By dynamically prioritizing updates from peers with aligned operational behaviors and preventing negative transfer from misaligned nodes, the global model learns a more robust boundary for normal behavior. Consequently, the system avoids the over-generalization typical of standard federated averaging, allowing it to correctly identify normal operating states across diverse environmental conditions without triggering false alarms. 

\subsection{Limitations}
While the proposed process mining-enabled federated anomaly detection approach demonstrates strong empirical performance, there are certain limitations to the current experimental setup that must be acknowledged.

First, the dataset utilized for evaluation was collected from a scale replica of an automotive manufacturing plant. Although this testbed effectively simulates various load and wear conditions across different operating modes, a scale model may not fully encapsulate the severe environmental noise, complex vibration interference, and compounding mechanical degradation present in a full-scale, operational industrial facility. Consequently, the generalizability of the similarity-weighted aggregation in highly chaotic, real-world environments requires further empirical validation.

Second, the ground truth for evaluating the anomaly detection models relies on synthetically injected anomalies. By shifting the normalized feature values by a random margin of 1.0 to 2.5 standard deviations, we established a mathematically rigorous baseline to test the boundary detection capabilities of the autoencoders. However, synthetic statistical shifts do not always perfectly mimic the nuanced, transient, and multi-modal mechanical faults that naturally occur in physical IoT equipment.

Third, while the comparison of our method with no federated learning involved and FedAvg establishes two reliable baselines that test whether our method addresses heterogeneity of the plant data and non-IID issues, it is worth noting that the literature includes many similarity techniques to address non-IID scenarios \cite{jimenezgutierrez2025fedlecc, zhang2024fedhc, fan2024rethinkingclusteredfl, dong2025fadngs, laridi2024enhanced}. A comprehensive comparison, including the analysis of control-flow aspects included in our method, would improve the overall experimentation strength.

Finally, while our method preserves communication efficiency by transmitting fitness scores instead of raw data during the similarity evaluation phase, it inherently introduces additional computational overhead at the edge. The local nodes are required to extract event logs via k-means clustering and perform conformance checking against the global process model. Depending on the hardware constraints of the deployed IoT sensors and gateways, this local computational burden may necessitate further algorithm optimization before broad deployment in resource-constrained edge networks.


\section{Conclusions}
\label{sec:conclusions}
This paper addressed federated anomaly detection in the presence of non-IID data, which is common in industrial IoT applications where multiple plants operate under different conditions. While existing federated anomaly detection approaches exist for non-IID data, they incur significant computational and communication overhead during the federated rounds and struggle to accurately capture the complex, sequential operational variations specific to industrial processes. In this paper, we propose a process mining-enabled, similarity-weighted federated anomaly detection approach that characterizes the dynamics of IoT plants and addresses non-IID conditions. Experimental results demonstrate that our proposal achieves superior performance compared to both isolated local training baselines and standard FedAvg. We plan on extending our work with actual embedded testbeds for practical evaluation of the feasibility of our proposal and perform a more comprehensive sensitivity analysis of the node similarity evaluation through process mining.

\bibliographystyle{IEEEtran}
\bibliography{bibliography}

\end{document}